%% file: samplepaper.tex
\documentclass[runningheads]{llncs}
\usepackage{graphicx,wrapfig,lipsum}
\usepackage{booktabs}
\usepackage{subcaption}
\usepackage{blindtext}
\usepackage[misc,geometry]{ifsym}
\usepackage[labelsep=period]{caption}
\usepackage[labelfont=bf]{caption}
\usepackage{tikz}
\usepackage{amsmath,amssymb}
\usepackage{textcomp}
\usepackage{soul,xcolor}
\newcommand\BigBox{\vcenter{\hbox{\scalebox{1.5}{$\Box$}}}}
\newcommand\bigsquare{\mathop{\BigBox}\limits}
 
\usepackage{multirow}

\AtBeginDocument{%
  \providecommand\BibTeX{{%
    \normalfont B\kern-0.5em{\scshape i\kern-0.25em b}\kern-0.8em\TeX}}}

\begin{document}
\setstcolor{red}
\title{Cross-modal and Cross-domain Knowledge Transfer for Label-free 3D Segmentation}
\titlerunning{Cross-modal and Cross-domain Knowledge Transfer}
%
\author{Jingyu Zhang${^*}$\inst{1} \and
Huitong Yang${^*}$\inst{2} \and
Dai-Jie Wu\inst{2} \and Jacky Keung\inst{1} \and
Xuesong Li\inst{4} \and Xinge Zhu\inst{3}$^{(\textrm{\Letter})}$ \and Yuexin Ma\inst{2}$^{(\textrm{\Letter})}$}
\authorrunning{J. Zhang \emph{et al}.}
%
%

\institute{City University of Hong Kong, Hong Kong SAR, China 
\and School of Information Science and Technology, ShanghaiTech University, China
\email{mayuexin@shanghaitech.edu.cn}
\and Chinese University of Hong Kong, Hong Kong SAR, China \\
\email{zhuxinge123@gmail.com}
\and College of Science, Australian National University, Canberra, Australia
}
\maketitle 
\begin{abstract}
Current state-of-the-art point cloud-based perception methods usually rely on large-scale labeled data, which requires expensive manual annotations. A natural option is to explore the unsupervised methodology for 3D perception tasks. However, such methods often face substantial performance-drop difficulties. Fortunately, we found that there exist amounts of image-based datasets and an alternative can be proposed, \emph{i.e.}, transferring the knowledge in the 2D images to 3D point clouds. Specifically, we propose a novel approach for the challenging cross-modal and cross-domain adaptation task by fully exploring the relationship between images and point clouds and designing effective feature alignment strategies. Without any 3D labels, our method achieves state-of-the-art performance for 3D point cloud semantic segmentation on SemanticKITTI by using the knowledge of KITTI360 and GTA5, compared to existing unsupervised and weakly-supervised baselines. 

{\let\thefootnote\relax\footnote{${^*}$ Equal contribution; $^{(\textrm{\Letter})}$Corresponding author}}

\end{abstract}
%
%

\input{introduction}

\input{related_work}

\input{methodology}

\input{experiments}

\input{conclusion}

\noindent\textbf{Acknowledgements}
This work was supported by NSFC (No.62206173), Natural Science Foundation of Shanghai (No.22dz1201900). 
%
%
%
\bibliographystyle{splncs04}
\bibliography{ref}

\end{document}

%% file: introduction.tex
Accurate depth information of large-scale scenes captured by LiDAR is the key to 3D perception. Consequently, LiDAR point cloud-based research has gained popularity. However, existing learning-based methods~\cite{zhu2021cylindrical,hou2022point,guo2020deep} demand extensive training data and significant labor power. Especially for the 3D segmentation task, people need to assign class labels to massive amounts of points. Such high cost really hinders the progress of related research. Therefore, relieving the annotation burden has become increasingly important for 3D perception.

Recently, plenty of label-efficient methods have been proposed for semantic segmentation. They try to simplify the labeling operation by using scene-level or sub-cloud-level weak labels~\cite{ren20213d}, or utilizing clicks or line segments as object-level or class-level weak labels~\cite{liu2021one,liu2022less}. 
However, these methods still could not get rid of the dependence on 3D annotations. Actually, annotating images is much easier due to the regular 2D representation. We aim to transfer the knowledge learned from 2D data with its annotations to solve 3D tasks. However, most related knowledge-transfer works~\cite{vu2019advent,huang2022category,peng2022cl3d} focus on the domain adaption under the same input modality, which is difficult to adapt to our task. Although recent works~\cite{sautier2022image,yan20222dpass} begin to consider distilling 2D priors to 3D space, they use synchronized and calibrated images and point clouds and leverage the physical projection between two data modalities for knowledge transfer, which limits the generalization capability.

\begin{figure}[t]
    \centering
    \includegraphics[scale=0.07]{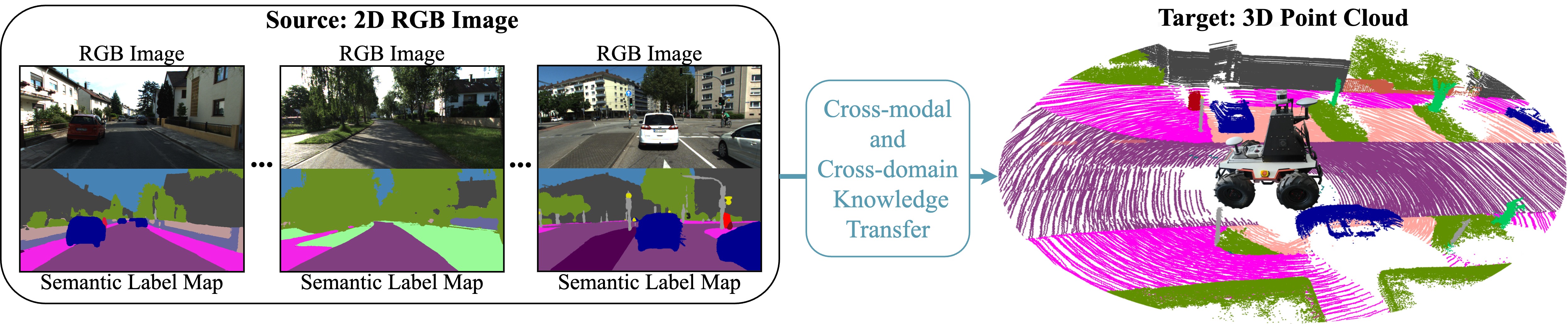}
    \caption{Our method performs 3D semantic segmentation by transferring knowledge from RGB images with semantic labels from arbitrary traffic datasets. }
    \label{fig:my_label}
\end{figure}

In this paper, we propose an effective solution for a novel and challenging task, i.e., cross-modal and cross-domain 3D semantic segmentation for LiDAR point cloud based on unpaired 2D RGB images with 2D semantic annotations from other datasets. 
However, images and point clouds have totally different representations and features, making it not trivial to directly transfer the priors. 
To deal with the large discrepancy between the two domains, we perform alignment at both the data level and feature level. For data level (i.e., data representation level), we transform LiDAR point clouds to range images by spherical projection to obtain a regular and dense 2D representation as RGB images. For the feature level, considering that the object distributions and relative relationships among objects are similar in traffic scenarios, we extract instance-wise relationships from each modality and align their features from both the global-scene view and local-instance view through GAN. In this way, our method makes 3D point cloud semantic features approach 2D image semantic features and further benefits the 2D-to-3D knowledge transfer. To prove the effectiveness of our method, we conduct extensive experiments on KITTI360-to-SemanticKITTI and GTA5-to-SemanticKITTI. Our method achieves state-of-the-art performance for 3D point cloud semantic segmentation without any 3D labels.

%% file: related_work.tex
\noindent \textbf{{LiDAR-based Semantic Segmentation:}}
Lidar-based 3D semantic segmentation \cite{milioto2019rangenet++,cortinhal2020salsanext} is crucial for autonomous driving. There are several effective representation strategies, such as 2D projection \cite{wu2018squeezeseg,gerdzhev2021tornado} and cylindrical voxelization \cite{zhu2021cylindrical}. However, most methods are in the full-supervision manner, which requires accurate manual annotations for every point, leading to high cost and time consumption. Recently, some research has been concentrated on pioneering labeling and training methods associated with weak supervision to mitigate the annotation workload. For instance, LESS \cite{liu2022less} developed a heuristic algorithm for point cloud pre-segmentation and proposed a label-efficient annotation policy based on pre-segmentation results. Scribble-annotation \cite{unal2022scribble} and label propagation strategies \cite{shi2022weakly} were also introduced. However, these methods still require expensive human annotation on the point cloud.
Several research studies \cite{jaritz2020xmuda,yan20222dpass,sautier2022image} have explored complementary information to enhance knowledge transfer from images and facilitate 2D label utilization. Our work discards the need for any 3D labels by transferring scene and semantic knowledge from 2D data to 3D tasks, which reduces annotation time and enhances practicality for real-world scenarios.

\noindent \textbf{{Unsupervised Domain Adaptation (UDA):}}
UDA techniques usually transfer the knowledge from the source domain with annotations to the unlabeled target domain by aligning target features with source features, which is most related to our task. For many existing UDA methods, the source and target inputs belong to the same modality, such as image-to-image and point cloud-to-point cloud domain adaptation. For image-based UDA, there are several popular methodologies for semantic segmentation, including adversarial feature learning \cite{huang2022category,vu2019advent,wu2021dannet}, pseudo-label self-training \cite{zhang2021prototypical,guo2022simt} and graph reasoning \cite{li2022sigma}. 
For point cloud-based UDA, it is more challenging due to the sparse and unordered point representation. Recent works~\cite{peng2022cl3d,bian2022unsupervised} have proposed using geometric transformations, feature space alignment, and consistency regularization to align the source and target point clouds. 
In addition, xMUDA \cite{jaritz2020xmuda} utilizes synchronized 2D images and 3D point clouds with 3D segmentation labels as the source dataset to assist 3D semantic segmentation. However, xMUDA assumes all modalities to be available during both training and testing, where we relax their assumptions by requiring no 3D labels, no synchronized multi-modal datasets, and during inference, no RGB data. Therefore, existing methods are difficult to extend to solve our cross-domain and cross-modality problem.
To the best of our knowledge, we are the first to explore the issue and propose an effective solution.

%% file: methodology.tex
\subsection{Problem Definition} \label{problem definition}
For the 3D semantic segmentation task, our model associates a class label with every point in a 3D point cloud $X \in R^{N \times 4}$, where $N$ denotes the  total number of points. Location information in 3D coordinates, and the reflectance intensity, are provided for each point, denoted by $\{x,y,z, i\}$. 
Without any 3D annotations, we conduct 3D segmentation by using only 2D semantic information on RGB images from arbitrary datasets in similar scenarios. Unlike existing UDA methods that perform adaptation under the same data modality, our task is more challenging because our source and target data are from different domains and different modalities, and can be named as \textit{Unsupervised Domain-Modality Adaptation (UDMA)}.
UDMA transfers knowledge from 2D source domain (images $I_S \in R^{H \times W \times 3}$ with semantic labels $Y_S \in R^{H \times W \times C}$, where $H$ and $W$ are height and width of the image, and $C$ is the number of classes) to 3D target domain (point clouds $X_T$). For UDMA 3D segmentation, we have $\{I_S, Y_S, X_T \}$ as the input data for training.  During inference, only 3D point clouds (projected to range images) are input to the model to produce 3D predictions of the shape $R^N$ (i.e., no RGB data are needed).



\begin{figure*}[t]
    \centering
    \includegraphics[scale=0.17]{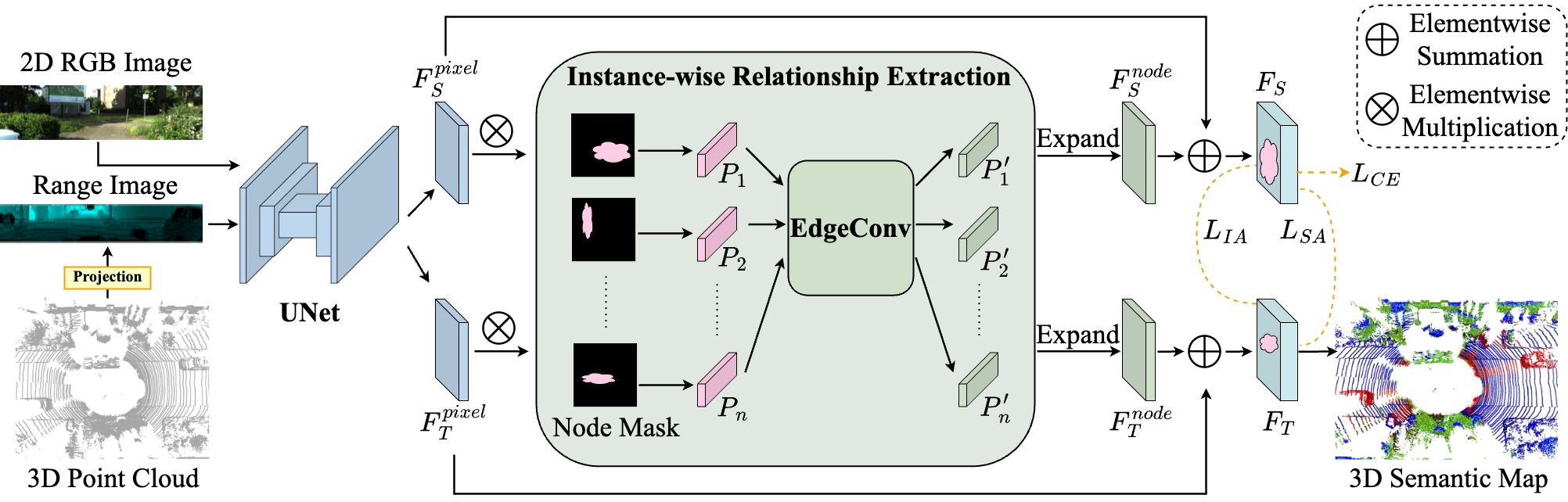}
    \caption{The pipeline of our method with feature extraction module, instance-wise relationship extraction module, and feature alignment module ($L_{IA}, L_{SA}$).}
    \label{fig:network}    
\end{figure*}

\subsection{Method Overview}
As shown in Fig. \ref{fig:network}, our model consists of three important modules: Feature Extraction, Instance-wise Relationship Extraction, and Feature Alignment.

\noindent\textbf{Feature Extraction:} This module is used to encode various input data to a certain feature representation. We apply UNet \cite{ronneberger2015u} as the feature extraction backbone, where the size of the output feature map retains the same as the input image size, which allows us to obtain pixel-level dense semantic features.

\noindent\textbf{Instance-wise Relationship Extraction (IRE):} This module carries the functionality of learning object-level features that encode local neighborhood information, therefore enhancing object-to-object interaction. The learned features are complementary to the mutually-independent pixel-level features learned by Feature Extraction module. Encoding more spatial information rather than domain- or modality-specific information enhances the efficiency of feature alignment. 

\noindent\textbf{Feature Alignment:} To solve UDMA tasks (defined in section \ref{problem definition}), we need to narrow down the 2D-3D domain gap to transfer label information. We achieve this through adversarial learning \cite{goodfellow2020generative} in both scene level and instance level, which learns to project inputs from different modalities to a common feature space and output domain- and modality-independent features. 






\begin{figure}[t]
    \centering
    \includegraphics[scale=0.075]{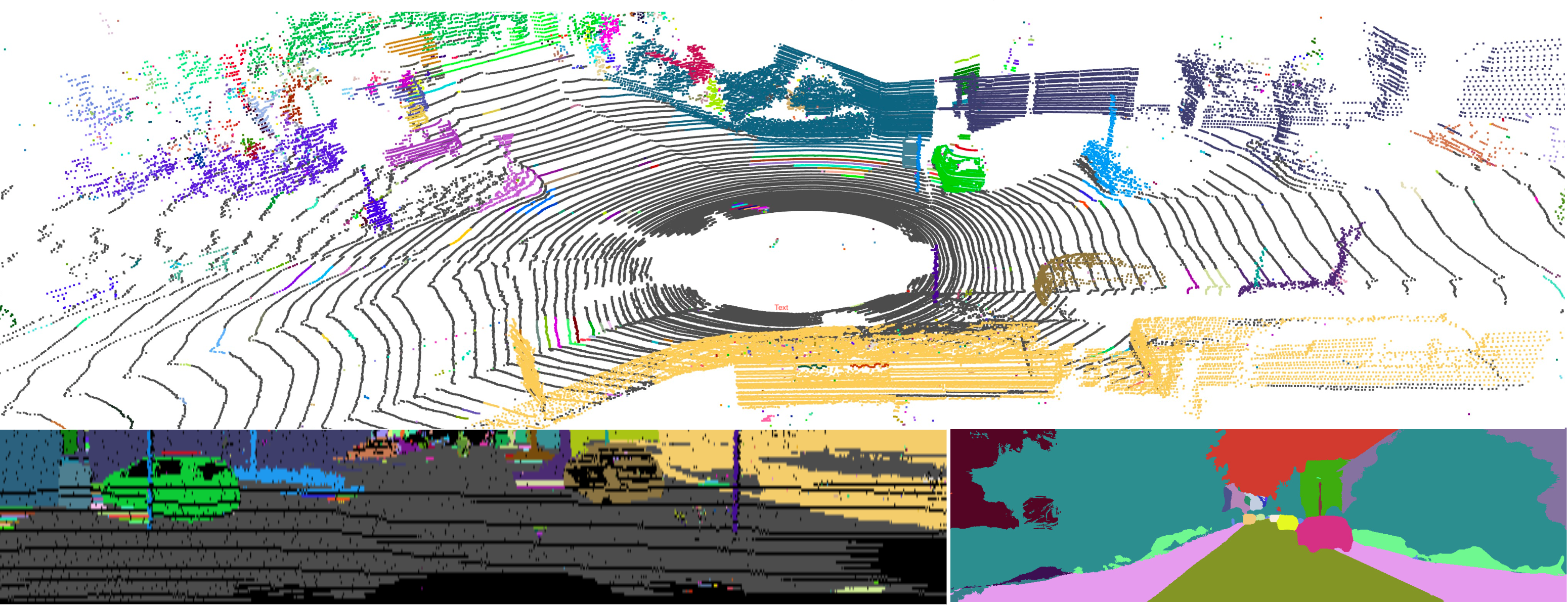}
    \caption{Visualization of Data-Pre-processing: Top: Point cloud pre-segmentation result. Bottom left: the range image after the spherical projection of the point cloud pre-segmentation. Bottom right: semantic labels in the RGB image.}
   
    \label{fig:pre_seg}
\end{figure}

\subsection{Data Pre-processing} \label{sec:data preprocess}

\noindent\textbf{LiDAR Point Cloud Pre-segmentation:}
Point clouds often contain valuable semantic information that can be used as prior knowledge without the need for any training. We utilize a pre-segmentation method proposed by LESS \cite{liu2022less} to group points into components such that each component is of high purity and generally contains only one object. The top image in Fig. \ref{fig:pre_seg} displays the pre-segmentation result. The approach involves using the RANSAC \cite{fischler1981random} for ground points detection. For non-ground points, an adaptive distance threshold is utilized to identify points belonging to the same component. We then utilize component information, such as mean or eigenvalues of point coordinates, to identify component classes. By using this heuristic approach we can successfully identify common classes (e.g., car, building, vegetation) with a high recall rate, and rare classes (e.g., person, bicycle) with a low recall rate. Due to the limited usefulness of low recall rate classes, we finally use three prior categories: car, ground (includes road, sidewalk, and terrain), and wall (building and vegetation) to assist the instance feature alignment in our method. In future work, we will explore the combination of learning-based and heuristic algorithms to improve the quantity and quality of the component class identification. For example, feature matching between unidentified and already identified components.



\noindent\textbf{Range Image Projection:}
2D and 3D data have huge a gap in their representations, e.g., structure, dimension, or compactness, which brings difficulties in knowledge transfer. To bridge the gap, we leverage another representation of 3D point cloud, range image \cite{langer2020domain,milioto2019rangenet++,wu2018squeezeseg}. 2D range images follow similar data representation as RGB (both represented in $height \times width$). As shown in the bottom two plots in Fig. \ref{fig:pre_seg}, the road (in gray and olive colors) is always located at the bottom of the picture with cars on top in both range and RGB images. The range image is generated by spherically projecting \cite{wu2018squeezeseg} 3D points onto a 2D plane:
\begin{equation}
\begin{aligned}
    u & = \frac{1}{2} \cdot (1 -arctan (y \cdot x^{-1} ) / \pi ) \cdot U \\
    v & = (1 - (\arcsin ( z \cdot r^{-1} ) +  f_{down} ) \cdot f^{-1}) \cdot V,
\end{aligned}
\end{equation}
where $u \in {1,...,U}$ and $v \in {1,...,V}$ indicate the 2D coordinate of the projected 3D point $(x,y,z)$. $r = \sqrt{x^2 + y^2 + z^2}$ is the range of each LiDAR point. We set $U=2048$ and $V=64$, respectively, which is ideal for Velodyne HDL-64E laser scan from SemanticKITTI dataset \cite{behley2019semantickitti}. 2D range images retain the depth information brought by LiDAR and provide rich image-level scene structure information. This helps in scene alignment. During training, we use $[r,r,i]$ for three input channels of range images, where $r$ and $i$ denote the range value defined above and the reflectance intensity, respectively.


\subsection{Instance-wise Relationship Extraction}
Based on the consideration that traffic scenarios share similar object location distribution and object-object interaction, we extract relationship features by proposing a novel graph architecture connecting instances as graph nodes. Therefore, it enhances the feature information with object-object interaction details and benefits the following feature alignment. Our Instance-wise Relationship Extraction (IRE) consists of two parts: Instance Node Construction and Instance-level Correlation.


\noindent\textbf{Instance Node Construction:} With the motivation of learning object-object relationships in an image, we regard every instance as a node in the input image instead of the whole class as a node.
For range images, we use pre-segmentation components as nodes, while for RGB images, we use ground truth to create node masks during training. Initial node descriptors $P_i$ are computed as follows:

\begin{equation}
\begin{aligned}
    Agg(F^{pixel} \otimes Mask_i) = P_i, \hspace{5mm} \forall i=1,...,n 
\end{aligned}
\end{equation}
The operation is the same for source and target data. $F^{pixel}$ represents the pixel-level dense feature map extracted by UNet, with the shape $R^{H \times W \times d}$ for source images ($F^{pixel}_S$) and $R^{V \times U \times d}$ for target images ($F^{pixel}_T$). Boolean matrices $Mask_i$ denote the binary node mask for the $i^{th}$ component, where the value $1$ denotes the pixel belonging to this component. The mask has the shape ${H \times W}$ in the source domain and ${V \times U}$ in the target domain. We denote $n$ by the total number of nodes in the current input image. With element-wise matrix multiplication operation $\otimes$, we obtain $m_i$ nonzero pixels features for each node, with shape $R^{m_i \times d}, i=1,...,n$. We apply an aggregation function $Agg$ to obtain a compact node descriptor $P_i \in R^d$. $Agg$ is composed of a pooling layer followed by a linear layer. Then, node descriptors are forwarded to EdgeConv to extract instance-level correlation.



\noindent\textbf{Instance-level Correlation:}
This submodule aims to densely connect the instance-level graph nodes and enhance recognizable location information for object interaction. After the Instance Node Construction, we embed local structural information into each node feature by learning a dynamic graph $G=(V, E)$ \cite{wang2019dynamic} in EdgeConv, where $V = \{ 1, ..., n \}$ and $E \subseteq V \times V$ are the vertices and edges. We build $G$ by connecting an edge between each node $P_i$ and its k-nearest neighbors (KNN) in $R^d$. 

\begin{equation}
    P'_i  = \mathop{\bigsquare}_{\{j | (i,j) \in E\}} f_{\Theta} (P_i, P_j)
\end{equation}
$f_{\Theta}$ learns edge features between nodes with learnable parameters $\Theta$. All local edge features of $P_i$ are then aggregated through an aggregation operation $\square$ (e.g., max or sum) to produce $P'_i \in R^d$ which is an enhanced descriptor for node $i$ that contains rich local information from neighbors. Instance-level correlation helps the model extract domain- and modality-independent features by focusing more on spatial-distribution features between instances. We then expand the instance features by replicating each $P'_i$ for $m_i$ times to obtain node-level feature map $F^{node}$ that is of the same size as $F^{pixel}$ (note that we omit the subscript here since the operation is the same for source and target data). We concatenate pixel-wise feature $F^{pixel}$ with instance-wise feature $F^{node}$ to form $F$, which keeps both dense individual information and local-neighborhood information to assist the feature alignment in the next section.

\subsection{Feature Alignment}
To transfer the knowledge obtained from RGB images to range images, we make the feature of range images approach that of RGB images from two aspects: the global scene-level alignment and local instance-level alignment. 

\noindent\textbf{Scene Feature Alignment (SA):}
Traffic scenarios in range and RGB images follow similar spatial distribution and object-object interaction (e.g., cars and buildings are located above the road). Therefore, we align the global scene features for two modalities.
Specifically, we implement the unsupervised alignment through Generative Adversarial Networks (GAN) \cite{goodfellow2020generative}. Adversarial training of the segmentation network can decrease the feature distribution gap between the source and target domain. Given the ground truth source labels, we train our model with the objective of producing source-like target features. In this way, we can maximize the utilization of low-cost RGB labels by transferring them to the target domain. This can be realized by two sub-process \cite{paul2020domain}: train the generator (i.e., the segmentation network) with target data to fool the discriminator; train the discriminator with both target and source data to discriminate features from two domains. In Scene Feature Alignment, we use one main discriminator to discriminate global features of the entire scene.

\noindent\textbf{Instance Feature Alignment (IA):}
Accurate classification for each point is critical for 3D semantic segmentation. We further design fine-grained instance-level feature alignment for two domains to improve the class recognition performance. To perform the instance-level alignment, we build extra category-wise discriminators to align source and target instances features. For target range images, as no labels are available, we use prior knowledge described in section \ref{sec:data preprocess} to filter prior categories, while for RGB we use ground truth. Then, we compute target features and align them with the corresponding source features of each prior category.

\subsection{Loss}
Our model is trained with three loss functions, with one supervised Cross-Entropy loss (CE loss) on source domain and two unsupervised feature alignment losses. Formally, we compute the CE loss for each pixel $(h,w)$ in the image and sum them over.

\begin{equation}
   L_{CE}(I_S,Y_S) = - \sum_{h} \sum_{w} \sum_{c} Y_S^{(h,w,c)} \log P_S^{(h,w,c)},
\label{eq:ce loss}
\end{equation}
where $h \in \{1,...,H\}$, $w \in \{1,...W\}$, and class $c \in \{ 1,...C\}$. $P_S$ is the prediction probability map of the source image $I_S$. For Scene Feature Alignment (SA), the adversarial loss is imposed on global target and source features $F_T, F_S$:

\begin{equation}
\begin{aligned}
    L_{SA} & (F_S, F_T; G, D) = - \log D (F_T) 
     - \log D(F_S) - \log (1-D(F_T)).
\label{Eq. scene loss}
\end{aligned}
\end{equation}

The output of discriminator $D(\cdot)$ is the domain classification, which is $0$ for target and $1$ for source. In SA, we use one main discriminator to discriminate range and RGB scene global features. For Instance Feature Alignment (IA), 
\begin{equation}
\begin{aligned}
     L_{IA} & (F^E_S, F^E_T; G, D^E) = \sum_{e=1}^{E} - y^e_T \log D^e (F^e_T) 
     - y^e_S \log D^e(F^e_S) - y^e_T \log (1-D^e(F^e_T)),
\label{Eq. instance loss}
\end{aligned}
\end{equation}

where $F^e_T$ and $F^e_S$ are the target and source features for each prior category $e$. $y^e_S = 1$ if category $e$ occurs in the current source image, otherwise $y^e_S = 0$, similarly for $y^e_T$.  Therefore, we only compute losses for the categories that are present in the current image. 

We use fine-tuning to further improve the performance. We apply weak label loss \cite{liu2022less} (Eq. \ref{eq: weak label loss}) for the prior categories that contain more than one class and CE loss $L_{CE}(X^e_T, Y^e_T)$ for the prior category that has fine-grained class identification, where $X^e_T \in R^{1 \times n_e \times 3}$ and $Y^e_T \in R^{1 \times n_e \times C}$. In our experiments, the weak label loss is applied to ground and wall, and the CE loss is applied to car,

\begin{equation}
    L_{weak} = - \frac{1}{n_e} \sum_{i=1}^{n_e} \log (1- \sum_{k_{ic} = 0} p_{ic}),
\label{eq: weak label loss}
\end{equation}
where $n_e$ indicates the number of pixels in the range image belonging to this prior category. $k_{ic} = 0$ if the class $c$ is not in this prior-category. $p_{ic}$ is the predicted probability of range pixel $i$ for class $c$. By using weak labels, we only punish negative predictions.

%% file: experiments.tex
\subsection{Datasets}
We evaluate our UDMA framework on two set-ups: \textit{KITTI360} $\rightarrow$ \textit{SemanticKITTI}  and \textit{GTA5} $\rightarrow$ \textit{SemanticKITTI}. KITTI360 \cite{liao2022kitti} contains more than 60k real-world RGB images with pixel-level semantic annotations of 37 classes. GTA5 \cite{Richter_2016_ECCV} is a synthetic dataset generated from modern commercial video games, which consists of around 25k images and pixel-wise annotations of 33 classes.
SemanticKITTI \cite{behley2019semantickitti} provides outdoor 3D point cloud data with $360^{\circ}$ horizontal field-of-view captured by 64-beam LiDAR as well as point-wise semantic annotation of 28 classes.


\subsection{Metrics and Implementation details}
To evaluate the semantic segmentation of point clouds, we apply the commonly adopted metric, mean intersection-over-union (mIoU) over $C$ classes of interests: 
\begin{equation}
    \frac{1}{C} \sum_{c=1}^{C} \frac{TP_c}{TP_c + FP_c + FN_c} ,
\end{equation}
where $TP_c$, $FP_c$, and $FN_c$ represent the number of true positive, false positive, and false negative predictions for class $c$. 
We report the labeling performance over six classes that frequently occur in driving scenarios on SemanticKITTI validation set.

\noindent\textbf{Implementation details:} 
We use the original size of KITTI360 and resize GTA5 images to [720,1280] for training. Our segmentation network is trained using SGD \cite{bottou2010large} optimizer with a learning rate $2.5 \times 10^{-4}$. We use Adam \cite{kingma2014adam} optimizer with learning rate $10^{-4}$ to train the discriminators which are fully connected networks with a hidden layer of size 256. 

\begin{table}[t]
\tiny
\caption{The first row presents the results of a modern segmentation backbone \cite{chen2017deeplab}. The medium five rows are 2D UDA methods \cite{vu2019advent,huang2022category,li2020content,tranheden2021dacs,wu2021dannet}. Ours is the result of our trained model with fine-tuning.}
\begin{subtable}[H]{0.45\textwidth}
\caption{KITTI360\ \textrightarrow \ SemanticKITTI.}
\centering
  \begin{tabular}{c c c cccccc }
    \hline
    Method & \rotatebox[origin=c]{90}{Road} & \rotatebox[origin=c]{90}{Sidewalk} & \rotatebox[origin=c]{90}{Building}  & \rotatebox[origin=c]{90}{Vegetation} & \rotatebox[origin=c]{90}{Terrain} &  \rotatebox[origin=c]{90}{Car} & mIoU \\
    \hline
    DeepLabV2 &17.96 & 6.79& 5.34& 33.24& 1.07& 0.71 &10.85 \\
    \hline
    AdvEnt & 20.34 & 10.87 & 14.48  & 34.13 & 0.06 & 0.21 & 13.35\\
    CaCo & 27.43 & 8.73 & 18.05  & 34.65 & 6.15 & 0.43 & 15.91 \\
    CCM & 0.01 & 5.63 & \textbf{24.15}   & 20.83 & 0 & 7.11  & 9.62 \\
    DACS & 19.35 & 7.44 & 0 & 0.13 & 0 & 6.64 & 5.59 \\
    DANNet & 15.30 & 0 & 13.75  & 31.46 & 0 & 0 & 10.09 \\
    \hline
    Ours  & \textbf{45.05} & \textbf{21.50} & 2.18 & \textbf{45.99} & \textbf{4.18}  & \textbf{28.49} & \textbf{24.57} \\
    \hline
\end{tabular}
\label{tab:360sem}
\end{subtable}
\hfill
 \begin{subtable}[ht]{0.475\textwidth}
 \caption{GTA5\ \textrightarrow  \ SemanticKITTI.}
\centering
  \begin{tabular}{c c c cccccc}
    \hline    
    Method & \rotatebox[origin=c]{90}{Road} & \rotatebox[origin=c]{90}{Sidewalk} & \rotatebox[origin=c]{90}{Building} & \rotatebox[origin=c]{90}{Vegetation} & \rotatebox[origin=c]{90}{Terrain} &  \rotatebox[origin=c]{90}{Car} & mIoU \\
    \hline
    DeepLabV2 & 18.87 & 0 & 24.19 & 17.46 & 0 & 0.97 & 10.24 \\
    \hline
    AdvEnt & 0.65 & 0 & 11.96  & 12.32 & 0 & 1.45 & 4.40 \\
    CaCo & 12.06 & 13.85 &  17.21  & 21.69 & 3.14 & 2.61 & 11.76 \\
    CCM & 0 & 2.18 & 20.47  & 20.14 & 0.27 & 0.83 & 7.32 \\
    DACS & 23.73 & 0 & 17.02   & 0 & 0 & 0 & 6.79 \\
    DANNet & 1.67 & \textbf{18.28} & \textbf{28.19} & 0 & \textbf{12.17} & 0 & 10.05 \\
    \hline
    Ours & \textbf{37.63} & 1.23 & 27.79 & \textbf{31.40} & 0.03 & \textbf{30.00} & \textbf{21.35} \\
  \hline
\end{tabular}
\label{tab:gta5sem}
\end{subtable}
\label{tab:main table}
\end{table}

\subsection{Comparison}
\noindent\textbf{Comparison with UDA methods:}
We evaluate the performance of adapting RGB images to range images under five 2D UDA state-of-the-art frameworks and one modern 2D semantic segmentation backbone (DeepLabV2). As shown in Table \ref{tab:main table}, our method outperforms all existing methods in terms of mIoU under both set-ups.
Compared to AdvEnt, CaCo, and DANNet which perform scene-level alignment through either adversarial or contrastive learning, our method achieves $8.66 \sim 14.48$ mIoU performance gain in KITTI360 and $9.59 \sim 16.95$ mIoU gain in GTA5 in terms of mIoU. This is attributed to our instance-level alignment and instance-wise relationship extraction modules. 
As for CCM and DACS, they show inferior results on both source datasets (even compared to other 2D UDA methods). 
DACS mixes images across domains to reduce the domain gap, which is more compatible when both domains follow the same color system. However, in our case, mixing $[R,G,B]$ pixels with $[r,r,i]$ range images is unreasonable and may distort the feature space. Similarly, the horizontal layout matrix used in CCM may fail in our setting as we have different horizontal field-of-view in range ($360^{\circ}$) and RGB images ($90^{\circ}$). They focus more on data-level modification. Our method accomplishes both data-level and feature-level alignment.
In our case, KITTI360 achieves higher mIoU, this may attribute to the fact that KITTI360 is using real-world images, whereas synthetic data in GTA5 may cause a large domain gap for learning.

\begin{table}[t]
\begin{minipage}{.6\linewidth}
\caption{Comparison with two 3D baselines.}
\centering
\scriptsize
\begin{tabular}{c c c ccccccccccc}
    \hline
    Method & \rotatebox[origin=c]{90}{Road} & \rotatebox[origin=c]{90}{Sidewalk} & \rotatebox[origin=c]{90}{Building}  & \rotatebox[origin=c]{90}{Vegetation} & \rotatebox[origin=c]{90}{Terrain} &  \rotatebox[origin=c]{90}{Car} & mIoU \\

    \hline
    Cylinder3D  & 7.63 & 5.18 & \textbf{9.30}  & 19.01 & \textbf{26.39} & 31.81 & 16.55 \\
    LESS  & 6.63 & 6.88 & 9.15 & 41.15 & 23.18  & \textbf{39.42} & 21.07 \\
    \hline
    Ours  & \textbf{45.05} & \textbf{21.50} & 2.18 & \textbf{45.99} & 4.18  & 28.49 & \textbf{24.57} \\   
    \hline
\end{tabular}
\label{tab:cylinder}
\end{minipage}%
\hspace*{5mm}
\begin{minipage}{.4\linewidth}
\vspace{-.5mm}
\scriptsize
\centering
\caption{Ablation Study:\\ effectiveness of IRE, SA, IA.}
\begin{tabular}{c c} 
\hline
 & \textbf{mIoU}\\
   \hline
      w/o IRE & 15.58 \\
      \hline
      w/o SA & 18.87 \\
      \hline
      w/o IA & 20.36 \\
      \hline
      \multirow{2}{2.6em}{Ours} & 22.53 \\
      & 24.57$^*$\\
        \hline
  \end{tabular}
\label{tab:ablationKITTI360}
\end{minipage}
\end{table}

\noindent\textbf{Comparison with 3D segmentation methods:}
We evaluate Cylinder3D \cite{zhu2021cylindrical} and LESS \cite{liu2022less} with prior labels obtained from pre-segmentation in Tab. \ref{tab:cylinder}. Cylinder3D is trained using CE loss for car and weak label loss for ground and wall. For LESS, we employ their proposed losses except ${L_{sparse}}$. Comparing with them, our method delivers a minimum of $3.50$ mIoU increase in performance. 3D baselines may produce suboptimal outcomes as they depend on the number of accurate 3D label annotations, while our approach utilizes rich RGB labels. We note that 3D-based methods provide better results in some classes (e.g., cars), which may be due to that they leverage 3D geometric features and representations more effectively, while our range image projection causes spatial information loss. Future work will explore the possibility of modifying our framework to function directly on the 3D point cloud without the need for projection.

\subsection{Ablation Studies}
Thorough experiments on KITTI360 $\rightarrow$ SemanticKITTI are conducted in Tab. \ref{tab:ablationKITTI360} for the effectiveness of network modules and fine-tuning. w/o denotes deleting the corresponding module in our network and $^*$ denotes adding fine-tuning.
We validate the effectiveness of \textit{IRE}, \textit{SA}, and \textit{IA} based on their $6.95$, $3.66$, and $2.17$ boost in mIoU, respectively (compared to 22.53). Through fine-tuning, we obtain another $+2.04$ in mIoU.

\begin{wrapfigure}[8]{r}{5cm}
\vspace{-20pt}
\includegraphics[width=5cm]{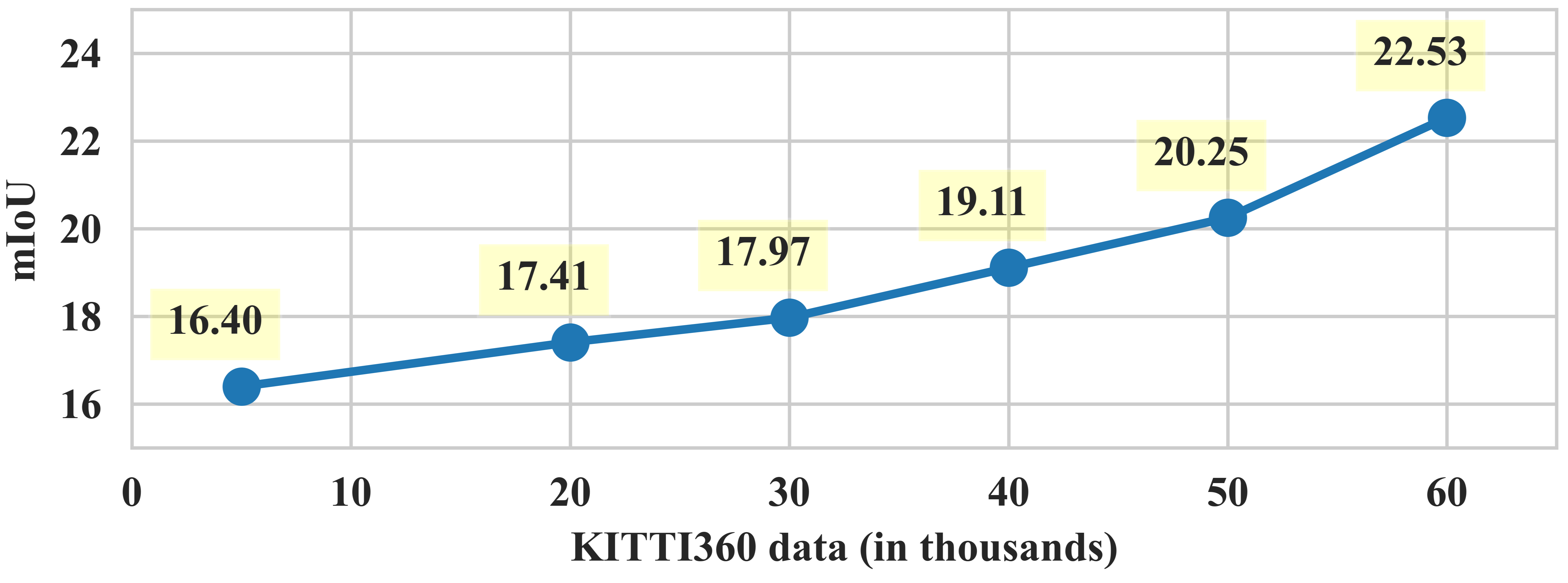}
  \caption{Analysis for different numbers of training data.}%
  \label{fig:line chart}  
\end{wrapfigure}
We also analyze how the number of training data in the source domain affects the result in the target domain in Fig. \ref{fig:line chart} under the setting KITTI360 $\rightarrow$ SemanticKITTI. 

The different number of training data is randomly sampled from KITTI360 with the same training protocol. The mIoU is lifted by $6.13$ with $55k$ more source data. More data allow the model to learn more comprehensive scene features capturing object interaction in driving scenarios. This demonstrates the potential efficacy of our proposed method, where further improvement in 3D mIoU is expected to be achieved with more annotated 2D training data.


%% file: conclusion.tex
In this work, we propose and address the cross-modal and cross-domain knowledge transfer task for 3D point cloud semantic segmentation from 2D labels to alleviate the burden of 3D annotations. We tackle the challenges via two complementary modules, Instance-wise Relationship Extraction and Feature Alignment. Extensive experiments illustrate the superiority and potential of our method.

%% file: samplepaper.bbl
\begin{thebibliography}{10}
\providecommand{\url}[1]{\texttt{#1}}
\providecommand{\urlprefix}{URL }
\providecommand{\doi}[1]{https://doi.org/#1}

\bibitem{behley2019semantickitti}
Behley, J., et~al.: Semantickitti: A dataset for semantic scene understanding
  of lidar sequences. In: ICCV. pp. 9297--9307 (2019)

\bibitem{bian2022unsupervised}
Bian, Y., et~al.: Unsupervised domain adaptation for point cloud semantic
  segmentation via graph matching. In: IROS. pp. 9899--9904. IEEE (2022)

\bibitem{bottou2010large}
Bottou, L.: Large-scale machine learning with stochastic gradient descent. In:
  COMPSTAT. pp. 177--186. Springer (2010)

\bibitem{chen2017deeplab}
Chen, L.C., et~al.: Deeplab: Semantic image segmentation with deep
  convolutional nets, atrous convolution, and fully connected crfs. TPAMI
  \textbf{40}(4),  834--848 (2017)

\bibitem{cortinhal2020salsanext}
Cortinhal, T., et~al.: Salsanext: Fast, uncertainty-aware semantic segmentation
  of lidar point clouds for autonomous driving. arXiv:2003.03653  (2020)

\bibitem{fischler1981random}
Fischler, M.A., et~al.: Random sample consensus: a paradigm for model fitting
  with applications to image analysis and automated cartography. CACM
  \textbf{24}(6),  381--395 (1981)

\bibitem{gerdzhev2021tornado}
Gerdzhev, M., et~al.: Tornado-net: multiview total variation semantic
  segmentation with diamond inception module. In: ICRA. pp. 9543--9549. IEEE
  (2021)

\bibitem{goodfellow2020generative}
Goodfellow, I., et~al.: Generative adversarial networks. CACM  \textbf{63}(11),
   139--144 (2020)

\bibitem{guo2022simt}
Guo, X., et~al.: Simt: handling open-set noise for domain adaptive semantic
  segmentation. In: CVPR. pp. 7032--7041 (2022)

\bibitem{guo2020deep}
Guo, Y., et~al.: Deep learning for 3d point clouds: A survey. TPAMI
  \textbf{43}(12),  4338--4364 (2020)

\bibitem{hou2022point}
Hou, Y., et~al.: Point-to-voxel knowledge distillation for lidar semantic
  segmentation. In: CVPR. pp. 8479--8488 (2022)

\bibitem{huang2022category}
Huang, J., et~al.: Category contrast for unsupervised domain adaptation in
  visual tasks. In: CVPR. pp. 1203--1214 (2022)

\bibitem{jaritz2020xmuda}
Jaritz, M., et~al.: xmuda: Cross-modal unsupervised domain adaptation for 3d
  semantic segmentation. In: CVPR. pp. 12605--12614 (2020)

\bibitem{kingma2014adam}
Kingma, D.P., et~al.: Adam: A method for stochastic optimization. arXiv
  preprint arXiv:1412.6980  (2014)

\bibitem{langer2020domain}
Langer, F., et~al.: Domain transfer for semantic segmentation of lidar data
  using deep neural networks. In: IROS. pp. 8263--8270. IEEE (2020)

\bibitem{li2020content}
Li, G., et~al.: Content-consistent matching for domain adaptive semantic
  segmentation. In: ECCV. pp. 440--456. Springer (2020)

\bibitem{li2022sigma}
Li, W., et~al.: Sigma: Semantic-complete graph matching for domain adaptive
  object detection. In: CVPR. pp. 5291--5300 (2022)

\bibitem{liao2022kitti}
Liao, Y., et~al.: Kitti-360: A novel dataset and benchmarks for urban scene
  understanding in 2d and 3d. TPAMI  (2022)

\bibitem{liu2022less}
Liu, M., et~al.: Less: Label-efficient semantic segmentation for lidar point
  clouds. In: ECCV. pp. 70--89. Springer (2022)

\bibitem{liu2021one}
Liu, Z., et~al.: One thing one click: A self-training approach for weakly
  supervised 3d semantic segmentation. In: CVPR. pp. 1726--1736 (2021)

\bibitem{milioto2019rangenet++}
Milioto, A., et~al.: Rangenet++: Fast and accurate lidar semantic segmentation.
  In: IROS. pp. 4213--4220. IEEE (2019)

\bibitem{paul2020domain}
Paul, S., et~al.: Domain adaptive semantic segmentation using weak labels. In:
  European conference on computer vision. pp. 571--587. Springer (2020)

\bibitem{peng2022cl3d}
Peng, X., et~al.: Cl3d: Unsupervised domain adaptation for cross-lidar 3d
  detection. arXiv:2212.00244  (2022)

\bibitem{ren20213d}
Ren, Z., et~al.: 3d spatial recognition without spatially labeled 3d. In: CVPR.
  pp. 13204--13213 (2021)

\bibitem{Richter_2016_ECCV}
Richter, S.R., et~al.: Playing for data: {G}round truth from computer games.
  In: ECCV. LNCS, vol.~9906, pp. 102--118. Springer (2016)

\bibitem{ronneberger2015u}
Ronneberger, O., et~al.: U-net: Convolutional networks for biomedical image
  segmentation. In: MICCAI. pp. 234--241. Springer (2015)

\bibitem{sautier2022image}
Sautier, C., et~al.: Image-to-lidar self-supervised distillation for autonomous
  driving data. In: CVPR. pp. 9891--9901 (2022)

\bibitem{shi2022weakly}
Shi, H., et~al.: Weakly supervised segmentation on outdoor 4d point clouds with
  temporal matching and spatial graph propagation. In: CVPR. pp. 11840--11849
  (2022)

\bibitem{tranheden2021dacs}
Tranheden, W., et~al.: Dacs: Domain adaptation via cross-domain mixed sampling.
  In: WACV. pp. 1379--1389 (2021)

\bibitem{unal2022scribble}
Unal, O., et~al.: Scribble-supervised lidar semantic segmentation. In: CVPR.
  pp. 2697--2707 (2022)

\bibitem{vu2019advent}
Vu, T.H., et~al.: Advent: Adversarial entropy minimization for domain
  adaptation in semantic segmentation. In: CVPR. pp. 2517--2526 (2019)

\bibitem{wang2019dynamic}
Wang, Y., et~al.: Dynamic graph cnn for learning on point clouds. TOG
  \textbf{38}(5),  1--12 (2019)

\bibitem{wu2018squeezeseg}
Wu, B., et~al.: Squeezeseg: Convolutional neural nets with recurrent crf for
  real-time road-object segmentation from 3d lidar point cloud. In: ICRA. pp.
  1887--1893. IEEE (2018)

\bibitem{wu2021dannet}
Wu, X., et~al.: Dannet: A one-stage domain adaptation network for unsupervised
  nighttime semantic segmentation. In: CVPR. pp. 15769--15778 (2021)

\bibitem{yan20222dpass}
Yan, X., et~al.: 2dpass: 2d priors assisted semantic segmentation on lidar
  point clouds. In: ECCV. pp. 677--695. Springer (2022)

\bibitem{zhang2021prototypical}
Zhang, P., et~al.: Prototypical pseudo label denoising and target structure
  learning for domain adaptive semantic segmentation. In: CVPR. pp.
  12414--12424 (2021)

\bibitem{zhu2021cylindrical}
Zhu, X., et~al.: Cylindrical and asymmetrical 3d convolution networks for lidar
  segmentation. In: CVPR. pp. 9939--9948 (2021)

\end{thebibliography}
